\documentclass[sigconf]{acmart}
\usepackage{underscore}
\usepackage{graphicx}
\usepackage{setspace}
\usepackage{algorithm}
\usepackage[noend]{algorithmic}

\AtBeginDocument{%
  \providecommand\BibTeX{{%
    \normalfont B\kern-0.5em{\scshape i\kern-0.25em b}\kern-0.8em\TeX}}}

\copyrightyear{2024}
\acmYear{2024}
\setcopyright{acmlicensed}\acmConference[MAD '24]{3rd ACM International
Workshop on Multimedia AI against Disinformation}{June 10, 2024}{Phuket,
Thailand}
\acmBooktitle{3rd ACM International Workshop on Multimedia AI against
Disinformation (MAD '24), June 10, 2024, Phuket, Thailand}
\acmDOI{10.1145/3643491.3660292}
\acmISBN{979-8-4007-0552-6/24/06}

\begin{document}

\title{Towards Quantitative Evaluation of Explainable AI Methods for Deepfake Detection}


\author{Konstantinos Tsigos}
\affiliation{%
  \institution{Information Technologies Institute}
  \city{CERTH, Thessaloniki}
  \country{Greece}
}
\email{ktsigos@iti.gr}

\author{Evlampios Apostolidis}
\affiliation{%
  \institution{Information Technologies Institute}
  \city{CERTH, Thessaloniki}
  \country{Greece}
}
\email{apostolid@iti.gr}

\author{Spyridon Baxevanakis}
\affiliation{%
  \institution{Information Technologies Institute}
  \city{CERTH, Thessaloniki}
  \country{Greece}
}
\email{spirosbax@iti.gr}

\author{Symeon Papadopoulos}
\affiliation{%
  \institution{Information Technologies Institute}
  \city{CERTH, Thessaloniki}
  \country{Greece}
}
\email{papadop@iti.gr}

\author{Vasileios Mezaris}
\affiliation{%
  \institution{Information Technologies Institute}
  \city{CERTH, Thessaloniki}
  \country{Greece}
}
\email{bmezaris@iti.gr}
\renewcommand{\shortauthors}{  }
\renewcommand{\shorttitle}{  }
\renewcommand\_{\textunderscore\allowbreak}

\begin{abstract}
In this paper we propose a new framework for evaluating the performance of explanation methods on the decisions of a deepfake detector. This framework assesses the ability of an explanation method to spot the regions of a fake image with the biggest influence on the decision of the deepfake detector, by examining the extent to which these regions can be modified through a set of adversarial attacks, in order to flip the detector's prediction or reduce its initial prediction; we anticipate a larger drop in deepfake detection accuracy and prediction, for methods that spot these regions more accurately. Based on this framework, we conduct a comparative study using a state-of-the-art model for deepfake detection that has been trained on the FaceForensics++ dataset, and five explanation methods from the literature. The findings of our quantitative and qualitative evaluations document the advanced performance of the LIME explanation method against the other compared ones, and indicate this method as the most appropriate  for explaining the decisions of the utilized deepfake detector.
\end{abstract}



\maketitle

\section{Introduction}

The recent advances in the field of Generative AI have led to new and more sophisticated ways of image and video manipulation, and the creation of a new type of visual disinformation that is often referred to as deepfakes. Deepfakes are AI manipulated media in which, a person's face or body is digitally altered in an existing image or video to make them appear as someone else or to reenact them. The ongoing improvement of Generative AI technologies enables the creation of deepfakes that are increasingly difficult to detect. The latter observation, combined with the use of deepfakes for spreading disinformation, necessitates the development of effective solutions for deepfake detection. Moreover, enhancing deepfake detection methods with explanatory mechanisms would significantly improve the users' trust in these technologies and allow obtaining insights about the applied image/video manipulation procedures for creating the detected deepfake.

Despite the growing interest in building increasingly more powerful models for deepfake detection, the provision of trustworthy explanations for the output of these models has not been studied extensively. Most works on explainable deepfake detection, investigate the use of various methods that create visual explanations (usually in the form of 2D heatmaps), but evaluate the performance of methods based only on the basis of qualitative analysis over a limited set of examples \cite{9092227,9707568,SILVA2022100217,9993294,10350382}. Only a recent work has attempted to assess the performance of various explanation methods on two CNN-based deepfake detection models using a quantitative evaluation framework \cite{GOWRISANKAR2024103684}. Nevertheless, their proposed framework uses explanations produced from correctly classified pristine (non-manipulated) images, in order to compare the performance of various explanation approaches. In contrast, we argue that the opposite use-case of explanations - i.e., when the model detects a deepfake - is both more meaningful and useful to the user. Moreover, their framework requires access to pairs of real-fake images, thus being non-applicable on datasets that contain only fake examples, e.g., the WildDeepfake dataset \cite{10.1145/3394171.3413769}.

In this paper, we propose a new evaluation framework that is simpler and more widely-applicable than the one in \cite{GOWRISANKAR2024103684}. The proposed framework takes into account the produced visual explanation for the deepfake detector's decision after correctly classifying a fake image, without requiring any access to its original counterpart. Based on this new framework, we evaluate the performance of five explanation methods from the literature on a state-of-the-art model for deepfake detection. Our contributions are the following:
\begin{itemize}
    \item We explain the decisions of a state-of-the-art model for deepfake detection, that is trained to spot four different types of deepfakes, i.e., deepfake attribution.
    \item We perform a comparative study among five different explanation methods, aiming to identify which is the most appropriate one for the considered model.
    \item We propose a new evaluation framework for quantifying the ability of explanation methods to spot the most influential image regions for the decision of a deepfake detection model. 
\end{itemize}

\section{Related Work}

Over the last years, there is an increasing interest in the development and training of advanced network architectures for deepfake detection. However, the explanation of the decisions of these networks has been poorly investigated. In an early work, Malolan et al. \cite{9092227}, trained a variant of the XceptionNet \cite{8099678} using a subset\footnote{Available at: https://github.com/ondyari/FaceForensics} of the FaceForensics++ dataset \cite{9010912} and examined the use of the LIME \cite{10.1145/2939672.2939778} and LRP \cite{10.1371/journal.pone.0130140} methods for producing visual explanations about the outcomes of the trained model. However, the evaluation of these methods was based on a few samples and mainly focused on the robustness of the produced explanations against various affine transformations or Gaussian blurring of the input image. Xu et al. \cite{9707568}, utilized the representations of EfficientNet-B0 \cite{DBLP:conf/icml/TanL19} and a supervised contrastive learning methodology to train a linear deepfake detector to discriminate the real from the manipulated images of the FaceForensics++ dataset \cite{9010912}. In terms of explainability, Xu et al. investigated the use of the learned features only for explaining the observed detection performance, using heatmap visualizations and uniform manifold approximation and projection (UMAP). Silva et al. \cite{SILVA2022100217}, proposed the use of an ensemble of CNNs (XceptionNet \cite{8099678}, EfficientNet-B3 \cite{DBLP:conf/icml/TanL19}) and attention-based models for deepfake detection. They provided explanations about the regions of the images that influence the most the decision of the detector, using the Grad-CAM method \cite{8237336} and focusing on the computed gradients for the attention map. Nevertheless, the produced explanations were evaluated only in a qualitative manner by taking into account only a few image samples. Jayakumar et al. \cite{9993294}, trained a deepfake detection model that utilizes the EfficientNet-B0 \cite{DBLP:conf/icml/TanL19} as backbone and contains five dense classification layers. To produce visual explanations, they investigated the use of the Anchors \cite{10.5555/3504035.3504222} and LIME \cite{10.1145/2939672.2939778} methods, and conducted evaluations based on a limited set of examples. Aghasanli et al. \cite{10350382}, described a deepfake detection model that relies on Vision Transformers and can be used for distinguishing original and fake images generated by various diffusion models. For explaining the model's output, Aghasanli et al. used SVM and xDNN \cite{ANGELOV2020185} classifiers to understand the model's behavior by analyzing the closest support vectors and prototypes for each classifier, respectively. The evaluation of the produced explanations though, was based on the qualitative analysis of few samples. Haq et al. \cite{10.1145/3624748} described a neurosymbolic deepfake detection method that is based on the idea that deepfakes exhibit inter- or intra- modality inconsistencies in the emotional expressions of the person being manipulated. Their method performs inter- and intra-modality reasoning on emotions extracted from audio and visual modalities using a psychological and arousal valence model for deepfake detection, and provides textual explanations that localize the timestamp and identify the fake part. However, it was evaluated only in terms of deepfake and emotion detection, while its explainability dimension was discussed only theoretically. Finally, Gowrisankar et al. \cite{GOWRISANKAR2024103684} described an evaluation framework for explanation methods, which is based on the intuition that the identified salient visual concepts by such a method after correctly classifying a real image as a non-manipulated one, could be used to flip the prediction of the detector for its fake counterpart. Initially, Gowrisankar et al., investigated the appropriateness of generic data removal/insertion approaches for modifying the spotted salient pixels or segments of the input image (e.g., zeroing, replacement with a uniform random value and blurring based on the neighboring pixels), and found out that these approaches may produce less meaningful results when applied on deepfake detection models, as they can distort facial regions and produce completely unexpected detection results (e.g., increase of deepfake detection accuracy). Based on this finding, they described a framework that applies a number of adversarial attacks (using Natural Evolution Strategies (NES) \cite{wierstra14a}) in regions of a fake image that correspond to the identified salient visual concepts after explaining the (correct) classification of its real counterpart, and evaluates the performance of an explanation method based on the observed drop in the accuracy of the deepfake detector. Thus, their evaluation framework takes the unusual step of using the produced explanation after correctly classifying a real (non-manipulated) image, in order to assess the capacity of an explanation method to explain the detection of a fake (manipulated) image.

\begin{figure*}[t]
\centering
\includegraphics[width=\textwidth]{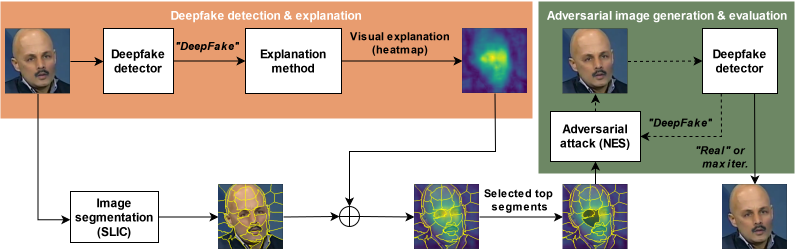}
\caption{The processing pipeline of the proposed evaluation framework.}
\label{fig:evalfram}
\end{figure*}

Differently to the majority of the works described above, that evaluate explanations qualitatively using a small set of samples \cite{9092227,9707568,SILVA2022100217,9993294,10350382}, in this work we assess the sufficiency of explanation methods to spot the regions of a manipulated image that influenced the most the deepfake detector's decision, using a quantitative evaluation framework. Our work is most closely related with \cite{GOWRISANKAR2024103684}, but we follow a more straightforward and intuitive evaluation approach that takes into account the produced explanations for the deepfake detector's output after correctly classifying a fake/manipulated image (which is better reflecting the task at hand). Moreover, we employ a state-of-the-art model for deepfake detection (rather than using out-of-date models, such as MesoNet \cite{DBLP:conf/wifs/AfcharNYE18} and XceptionNet \cite{8099678}), since there is no evidence from the literature that the results for one deepfake detector can be generalized to other detectors as well.

\section{Comparative Study Setup}

This section describes the employed deepfake detection model, explanation methods, and evaluation framework and measures.

\subsection{Deepfake detection model}

We use a model that relies on the second version of the EfficientNet architecture \cite{DBLP:conf/icml/TanL21} for deepfake detection. Building on the first version of EfficientNet - which leveraged Inverted Bottleneck convolutions (MBConv) and compound scaling to achieve high performance with fewer parameters compared to models with similar ImageNet accuracy \cite{DBLP:conf/icml/TanL19} - the employed version introduces Fused Inverted Bottleneck convolutions (Fused-MBConv), leading to even faster training and improved efficiency \cite{DBLP:conf/icml/TanL21}. We chose EfficientNet due to its widespread adoption, efficiency, and effectiveness in deepfake detection tasks, either as a part of an ensemble or as a backbone of more advanced methods \cite{DBLP:conf/cvpr/ShioharaY22, DBLP:conf/cvpr/ZhaoZ0WZY21}. Notably, an ensemble of five EfficientNet-B7 models achieved the winning performance in Meta's DFDC challenge \cite{DBLP:journals/corr/abs-1910-08854}. Moreover, EfficientNet has been shown to outperform alternative CNN architectures, such as XceptionNet \cite{8099678} and MesoNet \cite{DBLP:conf/wifs/AfcharNYE18} (that were taken into account in \cite{GOWRISANKAR2024103684}), on various deepfake datasets \cite{DBLP:journals/tcsv/LiNYFZ23, DBLP:journals/corr/abs-2310-07028, DBLP:conf/cvpr/HuangWYA00Y23}. Finally, it achieves similar performance to other vanilla CNNs on the ForgeryNet dataset \cite{DBLP:conf/cvpr/HeGCZYSSS021} while requiring fewer parameters.

\subsection{Explanation methods}

We produce visual explanations by highlighting the regions of the image (or video frame) with the biggest influence on the deepfake detection model's decision. As depicted in the orange coloured part of Fig. \ref{fig:evalfram}, we explain the outcome of the deepfake detector for a given fake image, using 2D heatmaps that represent the significance of different parts of the input image using a color scale. In our study, we consider the following explanation methods:
\begin{itemize}
    \item \textbf{Grad-CAM++} \cite{8354201}, is a back-propagation-based method that generates visual explanations by leveraging the information flow (gradients) during the back-propagation process. It extends the Grad-CAM method \cite{8237336}, by calculating a weighted combination of the positive partial derivatives of the last convolutional layer with respect to a specific class score in order to generate the visual explanation. In this way, it provides better (more complete) object localization and is capable of explaining occurrences of multiple instances of a given object in a single image.
    \item \textbf{RISE} \cite{DBLP:journals/corr/abs-1806-07421}, is a perturbation-based method that produces visual explanations by randomly masking out portions of the input image and assessing their impact on the model's output. Initially, this method generates a set of binary masks that are used to occlude regions of the input image and produce a set of perturbed images. Then, it feeds these perturbed images to the model, gets the model's predictions for each one of them and uses them to weight the corresponding binary masks. Finally, it creates the visual explanation by aggregating the weighted masks together.
    \item \textbf{SHAP} \cite{10.5555/3295222.3295230}, is an attribution-based method that leverages the Shapley values from game theory. It constructs an additive feature attribution model that attributes an effect to each input feature and sums the effects, i.e., SHAP values, as a local approximation of the output. More specifically, Shapley values assign importance scores to the individual pixels of the input image by treating them as players in a coalition game, with each player's presence or absence affecting the final outcome. The payout of the grand coalition is the prediction, or in our case the explanation, and Shapley values are used to divide this payout equally among pixels, by leveraging the model predictions for the perturbed images, to assess the contribution of each pixel to the prediction.
    \item \textbf{LIME} \cite{10.1145/2939672.2939778}, is a perturbation-based method that creates visual explanations by randomly masking out portions of an input image to assess their impact on the model's output. The fundamental idea behind LIME is the approximation of the model's behavior locally (i.e. around a specific instance) by generating a simpler, interpretable model. To this end, the input image is initially segmented and perturbed by randomly masking segments of it. Then, the perturbed images are given to the model that outputs its predictions. Finally, using a linear model (e.g., a linear regressor), LIME fits the binary masks of each perturbation to the corresponding scores and constructs the visual explanation by examining the coefficients/weights that emerge from this simpler model.
    \item \textbf{SOBOL} \cite{fel2021sobol}, is an attribution-based method that employs a mathematical concept called Sobol' indices (after Ilya M. Sobol’\footnote{https://en.wikipedia.org/wiki/Ilya_M._Sobol\%27}), to identify the contribution of the input variables on the variance of the model's output. Using a Quasi-Monte Carlo sequence, SOBOL generates a set of real-valued masks, which are then applied to an input image using perturbation functions such as blurring, to generate the perturbations. The resulting images are then forwarded to the model to get the prediction scores. By analyzing the relationship between the masks and their associated prediction scores, SOBOL estimates the total order of Sobol' indices and creates a visual explanation by highlighting the importance of each region.
\end{itemize}

\subsection{Evaluation framework and measures}

\begin{figure}[t]
\centering
\includegraphics[width=\columnwidth]{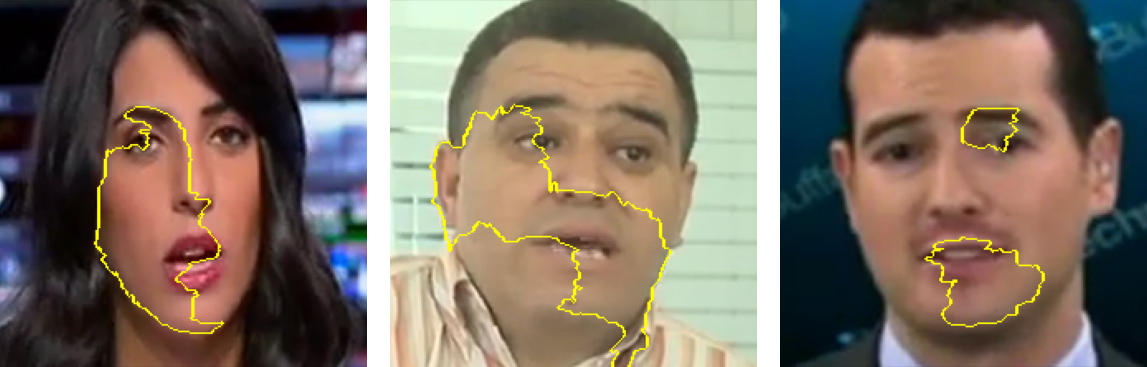}
\caption{The produced explanations by the LIME method (the best performing one according to the results in Section \ref{sec:experiments}), for three non-manipulated images of the FaceForensics++ dataset, that were correctly classified as ``real''.}
\label{fig:explanreal}
\end{figure}

Based on the reported findings in \cite{GOWRISANKAR2024103684}, about the adequacy of generic data removal/insertion approaches for perturbing the input image, we also do not apply such approaches on the image regions that have been promoted by an explanation method, in order to assess this method's performance. We evaluate the performance of an explanation method by extending the evaluation framework in \cite{GOWRISANKAR2024103684} so that it takes into account the produced explanations for fake images. We argue that the provision of an explanation after detecting a fake image is more meaningful for the user, as it can give clues about regions of the image (the highlighted ones by the visual explanation) that were found to be manipulated. On the contrary, the provided explanation after classifying an image as ``real'' would demarcate specific regions of the image as non-manipulated (see Fig. \ref{fig:explanreal}), while someone would expect that the entire image has not been manipulated at all.

Let us assume a fake image and the produced visual explanation for the deepfake detector's decision, by an explanation method (see the orange coloured part of Fig. \ref{fig:evalfram}). We assess the performance of this method by examining the extent to which the indicated regions in the visual explanation as the most important ones, can be used to flip the deepfake detector's decision (and thus classify the image as ``real''). For this, we segment the input image into super-pixel segments using the SLIC algorithm \cite{6205760}. Then, we quantify the contribution of each segment to the deepfake detector's decision by overlaying the created visual explanation to the segmented image and averaging the scores of the explanation for the pixels of the segment - as a note, in the case of LIME \cite{10.1145/2939672.2939778} we pass the SLIC-based segmentation mask of the input image and get the top-k scoring segments directly. Following, we focus on the top-k scoring segments and apply NES 
to progressively generate a variant of the input image that is classified as ``real'' by the deepfake detector. This iterative process, called adversarial image generation and evaluation, is illustrated in the green coloured area of Fig. \ref{fig:evalfram}. In each step of this process, we produce a variant of the input image by adding noise to the regions corresponding to the top-k scoring segments, following the steps in Alg. \ref{alg:advimgen}. The adversarial image generation and evaluation process stops if the deepfake detector classifies the adversarial image as ``real'' or a maximum number of iterations is reached.

\begin{algorithm}[t]
\caption{Adversarial image generation}\label{alg:advimgen}
\begin{algorithmic}[1]
\renewcommand{\algorithmicrequire}{\textbf{Parameters:}}
{\REQUIRE Search variance $\sigma$, Number of samples $\mathit{n}$, Image dimension $\mathit{N}$, Maximum number of iterations $\mathit{itr}$, Maximum distortion $\epsilon$, Learning rate $\alpha$}
\renewcommand{\algorithmicrequire}{\textbf{Input:}}
{\REQUIRE Deepfake image $\boldsymbol{x}$, Deepfake detector $\boldsymbol{F}$, Binary mask $\boldsymbol{M}$}
\renewcommand{\algorithmicrequire}{\textbf{Output:}}
{\ENSURE Adversarial image $x_{adv}$}
\setstretch{1.1}
\STATE $x_{adv} = x$,
\FOR{$i=1 \to itr$}
    \IF{$F(x_{adv})=real$}
        \STATE return $x_{adv}$
    \ENDIF
    \STATE $g = 0$
    \FOR{$j=1 \to n$}
        \STATE $u_{j} = N(0_{N},I_{N,N})$
        \STATE $g = g + F(x_{adv}[M] + \sigma u_{j}[M])_{label=real} \cdot u_{j}[M]$
        \STATE $g = g - F(x_{adv}[M] + \sigma u_{j}[M])_{label=real} \cdot u_{j}[M]$
    \ENDFOR
    \STATE $g = \frac{1}{2n\sigma}g$
    \STATE $x_{adv}[M] = x_{adv}[M] + clip_{\epsilon}(\alpha \cdot sign(g))$
\ENDFOR
\STATE return $x_{adv}$
\end{algorithmic}
\end{algorithm}

To quantify the performance of an explanation method, we calculate the accuracy of the deepfake detection model on the set of returned adversarial images after the completion of the adversarial image generation and evaluation process, when the adversarial attacks target the top-1, top-2 and top-3 scoring segments of the input images by the method. This measure ranges in $[0,1]$, where the upper boundary denotes a $100\%$ detection accuracy. We anticipate a larger decrease in accuracy for explanation methods that spot the most influential regions of the input image for the deepfake detector's decision, more effectively. Complementary to the aforementioned measure, we quantify the sufficiency of explanation methods to spot the most influential image regions for the deepfake detector, by calculating also the difference in the detector's output after applying adversarial attacks to the top-1, top-2 and top-3 scoring segments (following the paradigm in \cite{pmlr-v162-liu22i}). This measure ranges in $[0,1]$, where low/high sufficiency scores indicate that the top-k scoring segments by the explanation method have low/high impact to the deepfake detector's decision, and thus the produced visual explanation exhibits low/high sufficiency.

\section{Experiments}
\label{sec:experiments}
This section discusses the utilized dataset and implementation details, and reports the findings of our quantitative and qualitative evaluations. 

\begin{table*}[t]
\caption{The accuracy of the employed deepfake detection model for the different types of fakes in the FaceForensics++ dataset, on the original set of images (second row) and the adversarially-generated variants of them after modifying the image regions corresponding to the top-1, top-2 and top-3 scoring segments based on the different explanation methods. Best scores in bold and second best scores underlined.}
\centering
\begin{tabular}{|l|ccc|ccc|ccc|ccc|}
\hline
 &
  \multicolumn{3}{c|}{DF} &
  \multicolumn{3}{c|}{F2F} &
  \multicolumn{3}{c|}{FS} &
  \multicolumn{3}{c|}{NT} \\ \hline
Original Accuracy &
  \multicolumn{3}{c|}{0.978} &
  \multicolumn{3}{c|}{0.977} &
  \multicolumn{3}{c|}{0.982} &
  \multicolumn{3}{c|}{0.924} \\ \hline
 &
  \multicolumn{1}{c|}{Top 1} &
  \multicolumn{1}{c|}{Top 2} &
  Top 3 &
  \multicolumn{1}{c|}{Top 1} &
  \multicolumn{1}{c|}{Top 2} &
  Top 3 &
  \multicolumn{1}{c|}{Top 1} &
  \multicolumn{1}{c|}{Top 2} &
  Top 3 &
  \multicolumn{1}{c|}{Top 1} &
  \multicolumn{1}{c|}{Top 2} &
  Top 3 \\ \hline
Grad-CAM++ &
  \multicolumn{1}{c|}{0.781} &
  \multicolumn{1}{c|}{0.644} &
  0.571 &
  \multicolumn{1}{c|}{0.864} &
  \multicolumn{1}{c|}{0.798} &
  0.737 &
  \multicolumn{1}{c|}{0.887} &
  \multicolumn{1}{c|}{0.808} &
  0.728 &
  \multicolumn{1}{c|}{0.601} &
  \multicolumn{1}{c|}{0.481} &
  0.432 \\
RISE &
  \multicolumn{1}{c|}{0.877} &
  \multicolumn{1}{c|}{0.766} &
  0.686 &
  \multicolumn{1}{c|}{0.843} &
  \multicolumn{1}{c|}{0.710} &
  0.622 &
  \multicolumn{1}{c|}{0.896} &
  \multicolumn{1}{c|}{0.809} &
  0.734 &
  \multicolumn{1}{c|}{0.783} &
  \multicolumn{1}{c|}{0.637} &
  0.513 \\
SHAP &
  \multicolumn{1}{c|}{0.813} &
  \multicolumn{1}{c|}{0.609} &
  \underline{0.450} &
  \multicolumn{1}{c|}{0.846} &
  \multicolumn{1}{c|}{0.739} &
  0.637 &
  \multicolumn{1}{c|}{0.876} &
  \multicolumn{1}{c|}{\underline{0.702}} &
  \textbf{0.543} &
  \multicolumn{1}{c|}{0.686} &
  \multicolumn{1}{c|}{0.497} &
  0.344 \\
LIME &
  \multicolumn{1}{c|}{\textbf{0.735}} &
  \multicolumn{1}{c|}{\textbf{0.440}} &
  \textbf{0.245} &
  \multicolumn{1}{c|}{\textbf{0.803}} &
  \multicolumn{1}{c|}{\textbf{0.633}} &
  \textbf{0.484} &
  \multicolumn{1}{c|}{\textbf{0.864}} &
  \multicolumn{1}{c|}{\textbf{0.698}} &
  \underline{0.559} &
  \multicolumn{1}{c|}{\textbf{0.579}} &
  \multicolumn{1}{c|}{\textbf{0.340}} &
  \textbf{0.197} \\
SOBOL &
  \multicolumn{1}{c|}{\underline{0.750}} &
  \multicolumn{1}{c|}{\underline{0.591}} &
  0.490 &
  \multicolumn{1}{c|}{\underline{0.816}} &
  \multicolumn{1}{c|}{\underline{0.653}} &
  \underline{0.512} &
  \multicolumn{1}{c|}{\underline{0.874}} &
  \multicolumn{1}{c|}{0.703} &
  0.574 &
  \multicolumn{1}{c|}{\underline{0.621}} &
  \multicolumn{1}{c|}{\underline{0.417}} &
  \underline{0.313} \\ \hline
\end{tabular}
\label{tab:accuracy}
\end{table*}

\subsection{Dataset and implementation details}

Our experiments were conducted on the FaceForensics++ dataset \cite{9010912}. This dataset contains $1000$ original videos and $4000$ fake videos created using one of the following four classes of AI-based manipulation ($1000$ videos per class): ``FaceSwap'' (FS), ``DeepFakes'' (DF), ``Face2Face'' (F2F),  and ``NeuralTextures'' (NT). The videos of the FS class were created via a graphics-based approach that transfers the face region from a source to a target video. The videos of the DF class were produced using autoencoders to replace a face in a target sequence with a face in a source video or image collection. The videos of the F2F class were obtained by a facial reenactment system that transfers the expressions of a source to a target video while maintaining the identity of the target person. The videos of the NT class were generated by modifying the facial expressions corresponding to the mouth region, using a patch-based GAN-loss as utilized in Pix2Pix \cite{8100115}. The dataset is divided into training, validation, and test sets, comprised of $720$, $140$ and $140$ videos, respectively. 

For deepfake detection, we sampled the videos keeping 1 frame per second and used the RetinaFace face detector \cite{DBLP:conf/cvpr/DengGVKZ20} to obtain bounding boxes for the present faces. Following suggestions in \cite{9010912}, we enlarged each bounding box by a factor of $1.3$ to capture relevant background information that might aid in discriminating between real and fake samples. The cropped faces were stored and used as input to train and test the deepfake detector. For training, we leveraged a pre-trained model on the ImageNet 1K dataset obtained from the timm library \cite{rw2019timm}. Then, the deepfake detection model was trained for $30$ epochs using the AdamW optimizer \cite{DBLP:conf/iclr/LoshchilovH19} with a learning rate of $5\times 10^{-5}$ and a weight decay of $1\times 10^{-1}$, and the Cross-Entropy loss for multiclass classification. To mitigate overfitting and improve generalization, we employed the following data augmentation techniques: Random Erasing, Random Resized Crop, and AugMix \cite{DBLP:conf/iclr/HendrycksMCZGL20}. Additionally, to improve robustness to unseen data and encourage the model to learn more reliable features, we incorporated Stochastic Depth \cite{DBLP:conf/eccv/HuangSLSW16} with a drop path rate of $4\times 10^{-1}$. As a result, there was a $40\%$ chance of dropping a residual block connection during each forward pass.

To obtain the data for evaluating the different explanation methods we followed the approach in \cite{GOWRISANKAR2024103684}. In particular, we used $127$ videos from each different class of the test set and we sampled $10$ frames per video, thus creating four sets of $1270$ images. The generation of visual explanations was based on the following settings:
\begin{itemize}
    \item For \textbf{Grad-CAM++}, we took the average of all convolutional 2D layers.
    \item For \textbf{RISE}, we set the number of masks equal to $4000$ and kept all the other parameters with their default values.
    \item For \textbf{SHAP}, we set the number of evaluations equal to $2000$ and used a blurring mask with kernel size equal to $128$.
    \item For \textbf{LIME}, we set the number of perturbations equal to $2000$ and used the SLIC algorithm with a target number of segments equal to $50$.
    \item For \textbf{SOBOL}, we set the grid size equal to $8$ and the number of design equal to $32$, and kept all the other parameters with their default values.
\end{itemize}
With respect to NES, we set: the number of maximum iterations equal to $50$, the learning rate equal to $1/255$, the maximum distortion equal to $16/255$, the search variance equal to $0.001$, and the number of samples equal to $40$. All experiments were carried out on NVIDIA RTX 4090 GPU cards. The code for reproducing the reported results is publicly-available at: \url{https://github.com/IDT-ITI/XAI-Deepfakes}

\subsection{Quantitative results}

\begin{table*}[t]
\caption{The sufficiency scores of the considered explanation methods for the different types of fakes in the FaceForensics++ dataset, after modifying the top-1, top-2 and top-3 scoring segments of the input images. Best scores in bold and second best scores underlined.}
\centering
\begin{tabular}{|l|ccc|ccc|ccc|ccc|}
\hline
 &
  \multicolumn{3}{c|}{DF} &
  \multicolumn{3}{c|}{F2F} &
  \multicolumn{3}{c|}{FS} &
  \multicolumn{3}{c|}{NT} \\ \hline
 &
  \multicolumn{1}{c|}{Top 1} &
  \multicolumn{1}{c|}{Top 2} &
  Top 3 &
  \multicolumn{1}{c|}{Top 1} &
  \multicolumn{1}{c|}{Top 2} &
  Top 3 &
  \multicolumn{1}{c|}{Top 1} &
  \multicolumn{1}{c|}{Top 2} &
  Top 3 &
  \multicolumn{1}{c|}{Top 1} &
  \multicolumn{1}{c|}{Top 2} &
  Top 3 \\ \hline
Grad-CAM++ &
  \multicolumn{1}{c|}{0.148} &
  \multicolumn{1}{c|}{0.253} &
  0.310 &
  \multicolumn{1}{c|}{0.069} &
  \multicolumn{1}{c|}{0.115} &
  0.162 &
  \multicolumn{1}{c|}{0.063} &
  \multicolumn{1}{c|}{0.113} &
  0.160 &
  \multicolumn{1}{c|}{0.194} &
  \multicolumn{1}{c|}{0.251} &
  0.280 \\
RISE &
  \multicolumn{1}{c|}{0.087} &
  \multicolumn{1}{c|}{0.162} &
  0.219 &
  \multicolumn{1}{c|}{0.091} &
  \multicolumn{1}{c|}{0.173} &
  0.223 &
  \multicolumn{1}{c|}{0.060} &
  \multicolumn{1}{c|}{0.114} &
  0.157 &
  \multicolumn{1}{c|}{0.115} &
  \multicolumn{1}{c|}{0.204} &
  0.273 \\
SHAP &
  \multicolumn{1}{c|}{0.137} &
  \multicolumn{1}{c|}{0.300} &
  \underline{0.402} &
  \multicolumn{1}{c|}{0.092} &
  \multicolumn{1}{c|}{0.158} &
  0.222 &
  \multicolumn{1}{c|}{0.073} &
  \multicolumn{1}{c|}{\underline{0.181}} &
  \textbf{0.269} &
  \multicolumn{1}{c|}{0.167} &
  \multicolumn{1}{c|}{0.282} &
  0.357 \\
LIME &
  \multicolumn{1}{c|}{\textbf{0.195}} &
  \multicolumn{1}{c|}{\textbf{0.408}} &
  \textbf{0.539} &
  \multicolumn{1}{c|}{\textbf{0.121}} &
  \multicolumn{1}{c|}{\textbf{0.238}} &
  \textbf{0.334} &
  \multicolumn{1}{c|}{\textbf{0.087}} &
  \multicolumn{1}{c|}{\textbf{0.189}} &
  \underline{0.262} &
  \multicolumn{1}{c|}{\textbf{0.233}} &
  \multicolumn{1}{c|}{\textbf{0.363}} &
  \textbf{0.431} \\
SOBOL &
  \multicolumn{1}{c|}{\underline{0.166}} &
  \multicolumn{1}{c|}{\underline{0.277}} &
  0.352 &
  \multicolumn{1}{c|}{\underline{0.108}} &
  \multicolumn{1}{c|}{\underline{0.212}} &
  \underline{0.296} &
  \multicolumn{1}{c|}{\underline{0.078}} &
  \multicolumn{1}{c|}{0.180} &
  0.259 &
  \multicolumn{1}{c|}{\underline{0.198}} &
  \multicolumn{1}{c|}{\underline{0.302}} &
  \underline{0.362} \\ \hline
\end{tabular}
\label{tab:sufficiency}
\end{table*}

\begin{table*}[t]
\caption{Comparison of the obtained deepfake detection accuracy scores using our evaluation framework and the framework proposed in \cite{GOWRISANKAR2024103684}. Best scores in bold and second best scores underlined.}
\begin{tabular}{|l|cccc|cccc|}
\hline
                  & \multicolumn{4}{c|}{Our framework}                                                           & \multicolumn{4}{c|}{Framework in \cite{GOWRISANKAR2024103684}}                                                          \\ \hline
                  & \multicolumn{1}{c|}{DF}    & \multicolumn{1}{c|}{F2F}   & \multicolumn{1}{c|}{FS}    & NT    & \multicolumn{1}{c|}{DF}    & \multicolumn{1}{c|}{F2F}   & \multicolumn{1}{c|}{FS}    & NT    \\ \hline
Original Accuracy & \multicolumn{1}{c|}{0.930} & \multicolumn{1}{c|}{1.000} & \multicolumn{1}{c|}{1.000} & 0.973 & \multicolumn{1}{c|}{0.930} & \multicolumn{1}{c|}{1.000} & \multicolumn{1}{c|}{1.000} & 0.973 \\
Grad-CAM++        & \multicolumn{1}{c|}{0.462} & \multicolumn{1}{c|}{0.730} & \multicolumn{1}{c|}{0.823} & 0.544 & \multicolumn{1}{c|}{0.329} & \multicolumn{1}{c|}{0.486} & \multicolumn{1}{c|}{0.605} & 0.456 \\
RISE              & \multicolumn{1}{c|}{0.538} & \multicolumn{1}{c|}{0.527} & \multicolumn{1}{c|}{0.714} & 0.544 & \multicolumn{1}{c|}{0.285} & \multicolumn{1}{c|}{0.459} & \multicolumn{1}{c|}{0.585} & 0.456 \\
SHAP              & \multicolumn{1}{c|}{\underline{0.177}} & \multicolumn{1}{c|}{0.547} & \multicolumn{1}{c|}{\underline{0.558}} & 0.415 & \multicolumn{1}{c|}{\underline{0.247}} & \multicolumn{1}{c|}{\underline{0.378}} & \multicolumn{1}{c|}{\underline{0.537}} & \textbf{0.286} \\
LIME              & \multicolumn{1}{c|}{\textbf{0.101}} & \multicolumn{1}{c|}{\textbf{0.250}} & \multicolumn{1}{c|}{\textbf{0.476}} & \textbf{0.265} & \multicolumn{1}{c|}{0.367} & \multicolumn{1}{c|}{0.473} & \multicolumn{1}{c|}{0.599} & \underline{0.299} \\
SOBOL             & \multicolumn{1}{c|}{0.335} & \multicolumn{1}{c|}{\underline{0.324}} & \multicolumn{1}{c|}{0.571} & \underline{0.388} & \multicolumn{1}{c|}{\textbf{0.222}} & \multicolumn{1}{c|}{\textbf{0.331}} & \multicolumn{1}{c|}{\textbf{0.517}} & 0.354 \\ \hline
\end{tabular}
\label{tab:compacc}
\end{table*}

\begin{table*}[t]
\caption{Comparison of the obtained sufficiency scores using our evaluation framework and the framework proposed in \cite{GOWRISANKAR2024103684}. Best scores in bold and second best scores underlined.}
\begin{tabular}{|l|cccc|cccc|}
\hline
           & \multicolumn{4}{c|}{Our framework}                                                           & \multicolumn{4}{c|}{Framework in \cite{GOWRISANKAR2024103684}}                                                          \\ \hline
           & \multicolumn{1}{c|}{DF}    & \multicolumn{1}{c|}{F2F}   & \multicolumn{1}{c|}{FS}    & NT    & \multicolumn{1}{c|}{DF}    & \multicolumn{1}{c|}{F2F}   & \multicolumn{1}{c|}{FS}    & NT    \\ \hline
Grad-CAM++ & \multicolumn{1}{c|}{0.355} & \multicolumn{1}{c|}{0.211} & \multicolumn{1}{c|}{0.116} & 0.232 & \multicolumn{1}{c|}{0.397} & \multicolumn{1}{c|}{0.350} & \multicolumn{1}{c|}{0.252} & 0.319 \\
RISE       & \multicolumn{1}{c|}{0.280} & \multicolumn{1}{c|}{0.310} & \multicolumn{1}{c|}{0.173} & 0.276 & \multicolumn{1}{c|}{\underline{0.447}} & \multicolumn{1}{c|}{0.364} & \multicolumn{1}{c|}{0.262} & 0.317 \\
SHAP       & \multicolumn{1}{c|}{\underline{0.483}} & \multicolumn{1}{c|}{0.296} & \multicolumn{1}{c|}{\underline{0.276}} & \underline{0.356} & \multicolumn{1}{c|}{0.427} & \multicolumn{1}{c|}{\underline{0.382}} & \multicolumn{1}{c|}{\underline{0.279}} & \underline{0.411} \\
LIME       & \multicolumn{1}{c|}{\textbf{0.565}} & \multicolumn{1}{c|}{\textbf{0.495}} & \multicolumn{1}{c|}{\textbf{0.323}} & \textbf{0.441} & \multicolumn{1}{c|}{0.370} & \multicolumn{1}{c|}{0.331} & \multicolumn{1}{c|}{0.246} & \textbf{0.414} \\
SOBOL      & \multicolumn{1}{c|}{0.431} & \multicolumn{1}{c|}{\underline{0.434}} & \multicolumn{1}{c|}{\underline{0.276}} & 0.345 & \multicolumn{1}{c|}{\textbf{0.493}} & \multicolumn{1}{c|}{\textbf{0.444}} & \multicolumn{1}{c|}{\textbf{0.300}} & 0.376 \\ \hline
\end{tabular}
\label{tab:compsuff}
\end{table*}

Table \ref{tab:accuracy} reports the accuracy of the employed deepfake detection model for the different types of fakes in the FaceForensics++ dataset, on the original set of images (second row) and the adversarially-generated variants of them after modifying the image regions corresponding to the top-1, top-2 and top-3 scoring segments according to the different explanation methods. As shown in this table, the used deepfake detection model exhibits very high performance on all types of fakes of this dataset (achieving approx. $98\%$ accuracy on DF, F2F and FS and over $92\%$ on NT), documenting its state-of-the-art performance. With respect to the considered explanation methods, LIME appears to be the most effective one, as it is associated with the largest decrease in the detection accuracy for all types of fakes and in almost all experimental settings. As expected, the observed accuracy decrease is smaller when the adversarial image is generated based on the top-1 scoring segment and significantly larger when the adversarial attack is performed on the top-2 and top-3 scoring segments. However, this decrease is even more pronounced in the case of LIME. Therefore, LIME appears to be more effective compared to the other methods at highlighting the most influential segment of the input image for the decisions of the used deepfake detector, and noticeably better at spotting the top-2 or top-3 image segments with the highest impact on the detector's decision. Concerning the remaining methods, SOBOL seems to be the most competitive in most cases, while SHAP shows good performance in the case of DF and FS samples when spotting the top-2 or top-3 regions of the image. Finally, a comparison of the reported results across the different types of fakes, reveals that the different explanation methods can more effectively explain the detection of DF and NT classes, while the explanation of fakes from the remaining two classes is a more challenging task.

Table \ref{tab:sufficiency} presents the sufficiency scores of the considered explanation methods for the different types of fakes in the FaceForensics++ dataset, after performing adversarial attacks at the top-1, top-2 and top-3 scoring segments of the input images. These scores seem to be aligned with the results in Table \ref{tab:accuracy}, demonstrating once again, that LIME performs consistently good for all the considered types of fakes and numbers of top-scoring segments. Moreover, its effectiveness in spotting the most influential regions of the images is more pronounced when taking into account the top-3 scoring segments according to the produced visual explanation. As before, SOBOL is the second best method and SHAP performs comparatively good in specific occasions. Finally, the most challenging cases in terms of visual explanation, still remain the ones associated with fakes of the F2F and FS classes.

Finally, we compared the obtained results after applying the proposed evaluation framework and the one in \cite{GOWRISANKAR2024103684}. As a note, this comparison was based on a subset of (randomly) selected images ($150$ per class of fakes) to limit the computational needs of the experiment. The scores about the deepfake detector's accuracy and the explanation method's sufficiency are reported in Tables \ref{tab:compacc} and \ref{tab:compsuff}, respectively. These demonstrate that the different frameworks lead to different outcomes about the performance and the ranking of the considered explanation methods. Once again, LIME is the best-performing method for the selected subset of images according to our framework, while the framework from \cite{GOWRISANKAR2024103684} points to SOBOL as the most effective method. The observed difference is explained by the fact that the two frameworks base their evaluations on different conditions. The framework from \cite{GOWRISANKAR2024103684} assesses the performance of an explanation method by taking into account the produced explanations for correctly classified non-manipulated images. On the contrary, our framework focuses on the produced explanations for manipulated images that were classified as deepfakes, since highlighting the image regions that were perceived as manipulated is more meaningful for the users.

\begin{figure*}[!t]
  \centering
  \includegraphics[width=0.92\textwidth,height=0.92\textheight,keepaspectratio]{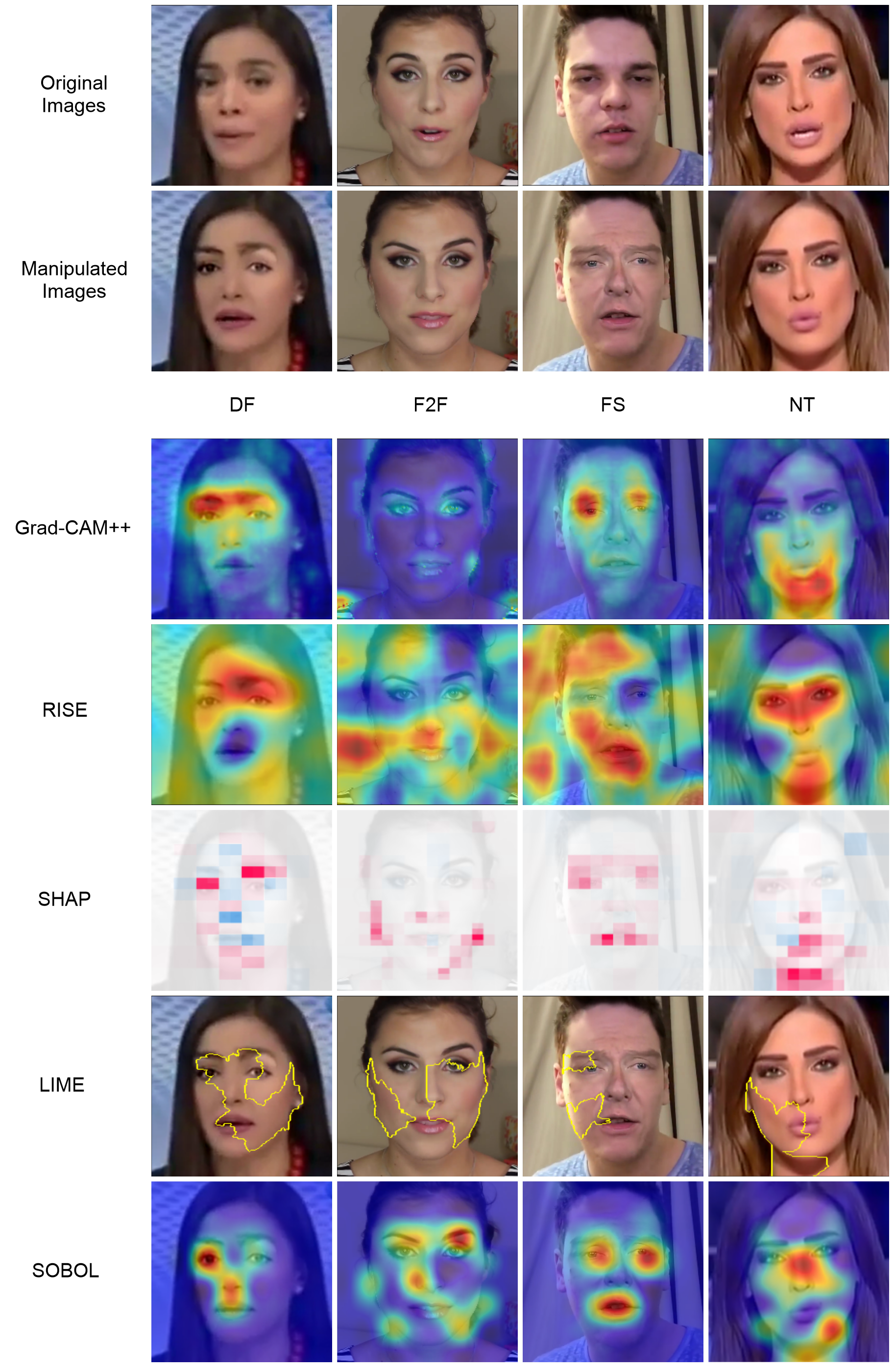}
  \caption{The obtained visual explanations from the considered explanation methods for four different images of the FaceForensics++ dataset (one per different type of manipulation). In terms of visualization, we adopt the default supported format by each explanation method.}
  \label{fig:qual}
\end{figure*}

\subsection{Qualitative results}

The top row of Fig. \ref{fig:qual} shows four different images (sampled video frames) of the FaceForensics++ dataset and the next row contains their AI manipulated variants, where each variant is associated with a different type of manipulation. The remaining rows present the produced visual explanations by the examined methods. As illustrated in these rows, LIME successfully spots: i) the regions close to the eyes and mouth that have been modified in the case of the DF sample, ii) the regions around the nose and the cheeks that have been changed in the case of the F2F sample, iii) the regions close to the left eye and cheek that have been altered in the case of the FS sample, and iv) the regions close to the mouth and chin that have been manipulated in the case of the NT sample. With respect to the other explanation methods, Grad-CAM++ correctly focuses on regions close to the eyes in the DF and FS samples and close to the chin in the case of the NT sample. However, it fails to clearly indicate regions in the case of the F2F sample and to spot manipulations around the mouth in the case of the DF sample. RISE seems to produce explanations that highlight irrelevant (see the F2F and FS samples) or non-manipulated regions (see the NT sample) of the image, while also failing to spot the manipulated ones (see the DF sample). Finally, SHAP and SOBOL appear to perform well compared with LIME, as in most cases, they provide explanations that indicate the altered regions of the images. This finding is aligned with the performance of these methods according to the conducted quantitative evaluation.  

\section{Conclusions}

In this paper, we presented a new evaluation framework for explainable AI methods for deepfake detection, which measures the capacity of such methods to spot the most influential regions of the input image through an adversarial image generation and evaluation process that aims to flip the detector's decision. We applied this framework on a state-of-the-art model for deepfake detection and five explanation methods from the literature. Our experimental results demonstrate the competitive performance of the LIME explanation method across all different types of fakes, and its competency to produce meaningful explanations for the employed deepfake detection model.

\begin{acks}
This work was supported by the EU Horizon Europe and Horizon 2020 programmes under grant agreements 101070190 AI4TRUST and 951911 AI4Media, respectively.
\end{acks}



\end{document}